\documentclass[letterpaper, 10 pt, conference]{ieeeconf}

\IEEEoverridecommandlockouts

\usepackage{cite}
\usepackage{amsmath,amssymb,amsfonts}
\usepackage{algorithmic}
\usepackage{graphicx}
\usepackage{textcomp}
\usepackage{xcolor}
\usepackage{siunitx}
\usepackage{float}
\usepackage{ar}
\usepackage{balance}

\newcommand{\bs}[1]{\boldsymbol{#1}}  
\newcommand{\ts}[1]{\text{#1}}

\renewcommand{\baselinestretch}{0.842}

\def\BibTeX{{\rm B\kern-.05em{\sc i\kern-.025em b}\kern-.08em

    T\kern-.1667em\lower.7ex\hbox{E}\kern-.125emX}}

\begin{document}

\title{\LARGE \bf Force Characterization of Insect-Scale Aquatic Propulsion Based on Fluid--Structure Interaction

\thanks{The research presented in this paper was partially funded by the Washington State University (WSU) Foundation and the Palouse Club through a Cougar Cage Award to \mbox{N.\,O.\,P.-A.}; the US National Science Foundation (NSF) through \mbox{Award\,\,2244082}; the Morgan Family Charitable Trust through a direct gift to \mbox{N.\,O.\,P.-A.}; and, the Koerner Family Foundation (KFF) through a graduate fellowship to \mbox{C.\,K.\,T.}. Additional funding was provided by the WSU Voiland College of Engineering and Architecture through a \mbox{start-up} fund to \mbox{N.\,O.\,P.-A.}.} %
\thanks{The authors are with the School of Mechanical and Materials Engineering, Washington State University (WSU), Pullman,\,WA\,99164,\,USA. Corresponding authors' \mbox{e-mail:} {\tt conor.trygstad@wsu.edu}~(C.\,K.\,T.); {\tt n.perezarancibia@wsu.edu} (N.\,O.\,P.-A.).}%
}
\author{Conor K. Trygstad and N\'estor O. P\'erez-Arancibia}

\maketitle
\thispagestyle{empty}
\pagestyle{empty}

\begin{abstract}
We present force characterizations of two newly developed \mbox{insect-scale} propulsors---one single-tailed and one double-tailed---for microrobotic swimmers that leverage \textit{fluid--structure interaction} (FSI) to generate thrust. The designs of these two devices were inspired by anguilliform swimming and are driven by soft tails excited by \textit{\mbox{high-work-density}}~(HWD) actuators powered by \textit{shape-memory alloy}~(SMA) wires. While these propulsors have been demonstrated to be suitable for microrobotic aquatic locomotion and controllable with simple architectures for trajectory tracking in the \emph{two-dimensional}~($\bs{2}$D) space, the characteristics and magnitudes of the associated forces have not been studied systematically. In the research presented here, we adopted a theoretical framework based on the notion of reactive forces and obtained experimental data for characterization using a \mbox{custom-built} \mbox{\textmu{N}-resolution} force sensor. We measured maximum and \mbox{cycle-averaged} force values with \mbox{multi-test} means of respectively \mbox{$\bs{0.45}$\,mN} and \mbox{$\bs{2.97}\,${\textmu}N}, for the tested \mbox{single-tail} propulsor. For the \mbox{dual-tail} propulsor, we measured maximum and \mbox{cycle-averaged} force values with \mbox{multi-test} means of \mbox{$\bs{0.61}$\,mN} and \mbox{$\bs{22.6}\,$\textmu{N}}, respectively. These results represent the first measurements of the instantaneous thrust generated by \mbox{insect-scale} propulsors of this type and provide insights into FSI for efficient microrobotic propulsion.
\end{abstract}

\section{Introduction} 
\begin{figure}[t!]
\vspace{1.5ex}
\begin{center}
\includegraphics[width=0.48\textwidth]
{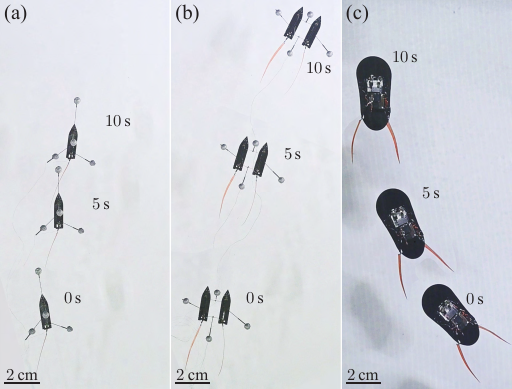}
\end{center}
\vspace{-2ex}
\caption{\hspace{-1ex}\textbf{Image composites of frames, taken at \mbox{$\boldsymbol{5}$-s} intervals, showing three swimming tests performed using three different \mbox{insect-scale} microrobots driven by \mbox{FSI-based} aquatic propulsors.}~\textbf{(a)}\,The \mbox{single-tail} \mbox{$45$-mg} VLEIBot swimmer presented in\cite{VLEIBot_2024}, while operating excited by a \mbox{$1$-Hz} PWM signal with a DC of \mbox{$8$\hspace{0.1ex}\%}.~\textbf{(b)}\,The \mbox{dual-tail} \mbox{$90$-mg} \mbox{VLEIBot\textsuperscript{+}} swimmer presented in\cite{VLEIBot_2024}, while operating excited by a \mbox{$1$-Hz} PWM signal with a DC of \mbox{$8$\hspace{0.1ex}\%}.\,\textbf{(c)}\,The \mbox{dual-tail} \mbox{$900$-mg} autonomous \mbox{VLEIBot\textsuperscript{++}} swimmer presented in\cite{LongwellCR2024}, while operating excited by a \mbox{$1$-Hz} PWM signal with a DC of \mbox{$5$\hspace{0.1ex}\%}. These experiments can be viewed in the attached supplementary movie. \label{FIG01}}
\vspace{-2.0ex}
\end{figure}

\vspace{-0.5ex}
\label{SECTION01}
Multimodal microrobots capable of working collaboratively in large autonomous teams have the potential to enhance our technological capabilities in complex environments, such as agricultural fields, confined spaces inside \mbox{earthquake-collapsed} buildings, narrow cavities within coral reefs, and deep waters. During the past few years, the robotics community has made significant technological progress in aquatic\cite{VLEIBot_2024,LongwellCR2024,Spino2024SwimmingBomb,Berlinger2021AutonomousFish,Berlinger2018DEAFish,Marchese2014AutonomousSoftFish,TrygstadCK2024,Zhao_OrigamiSwimmer_2022,Zhao2021PZTFrog,Waterstrider_2023}, terrestrial\cite{goldberg2018power,Ji2019DEAnsect,RoBeetle_2020,SMALLBug_2020,MilliMobile2023,wu2019insect, gravish2020stcrawler,SMARTI_2021,Waterstrider_2023}, and aerial microrobotics\cite{BeePlus_2019, BeePlusPlus_2023, ren2022DEAFlyer, Fuller2017Dampers, Chen2015AerialAquaticRobobee, Chen2017HybridBioinspiredVehicle}. However, fundamental challenges in energy storage, manufacturing, control, actuation, sensing, and embedded intelligence must be addressed before the vision of autonomous collaborative microrobotics can become a reality. One major remaining challenge for \mbox{mm-to-cm--scale} locomoting robots is the ability to generate large forces relative to their size and weight to achieve high maneuverability. Regarding \mbox{cm-scale} aquatic propulsion, we have recently developed two new devices that utilize soft tails to produce thrust through \textit{fluid--structure interaction}~(FSI), driven by actuators powered by \textit{shape-memory alloy}~(SMA) wires. Namely, the \mbox{single-tail} propulsor created to drive the \mbox{$45$-mg} VLEIBot swimmer presented in\cite{VLEIBot_2024} and shown in~\mbox{Fig.\,\ref{FIG01}(a)}, and the \mbox{dual-tail} propulsor developed to achieve the \textit{two-dimensional} \mbox{($2$D)} controllability on the surface of water of the \mbox{$90$-mg} VLEIBot\textsuperscript{+} swimmer also presented in\cite{VLEIBot_2024} and shown in~\mbox{Fig.\,\ref{FIG01}(b)}. Together, these innovations enabled the creation of the \mbox{$900$-mg} autonomous VLEIBot\textsuperscript{++} swimmer shown in \mbox{Fig.\,\ref{FIG01}(c)\cite{LongwellCR2024}}. Although the functionality of these \mbox{insect-scale} thrusters has been empirically demonstrated, the characteristics of the forces produced by the two different designs have not yet been investigated.

Here, we present a combined  theoretical and experimental framework to characterize the forces produced by the two propulsors described above. Specifically, we consider a model for the soft tail driving a propulsor using reactive theory\cite{LighthillMJ1971}, which provides valuable insights for design, and for parameter analysis and selection. Also, we compare predictions obtained through this model with measurements collected employing a \mbox{mm-scale} \mbox{{\textmu}N-resolution} force sensor designed and built using the techniques presented in\cite{Singer2019ClipBrazing} and the references therein. With this sensing device, for the tested \mbox{single-tail} propulsor operating at a frequency of \mbox{$1$\,Hz}, we measured maximum and \mbox{cycle-averaged} force values with \mbox{multi-test} means of \mbox{$0.45$\,mN} and \mbox{$2.97$\,{\textmu}N}, respectively. For the \mbox{dual-tail} propulsor operating at a frequency of \mbox{$1$\,Hz}, we measured maximum and \mbox{cycle-averaged} force values with \mbox{multi-test} means of \mbox{$0.61$\,mN} and \mbox{$22.6$\,{\textmu}N}, respectively. Comparisons of the experimental data with the results obtained through the \mbox{reactive-force} model indicate that this mathematical description is reasonably accurate in predicting the thrust generated by a \mbox{single-tail} propulsor, while failing at predicting the force produced by the \mbox{dual-tail} device. These observations suggest that the derived reactive model of force generation is applicable to situations in which a slender single tail interacts with a surrounding fluid in isolation. However, it fails at capturing the generation of hydrodynamic forces produced by the interaction of wake structures associated with two or more propulsive tails. We hypothesize that this issue can be remedied by incorporating a convective term into the model, which in this study we neglected \textit{a~priori} because during the performance of force tests, the tail does not move forward relative to the inertial frame of reference.    

The analyses and experimental results presented in this paper represent the first systematic study of the mechanism of hydrodynamic force generation through FSI phenomena at the insect scale. The rest of the paper is organized as follows. \mbox{Section\,\ref{SECTION02}} presents the derivation of the proposed reactive model of force generation. \mbox{Section\,\ref{SECTION03}} describes the design and fabrication of the \mbox{{\textmu}N-resolution} sensor used during force characterization. \mbox{Section\,\ref{SECTION04}} presents the collected experimental data. Last, \mbox{Section\,\ref{SECTION05}} summarizes the presented research.
\begin{figure}
\vspace{1.5ex}
\begin{center}    
\includegraphics[width=\linewidth]{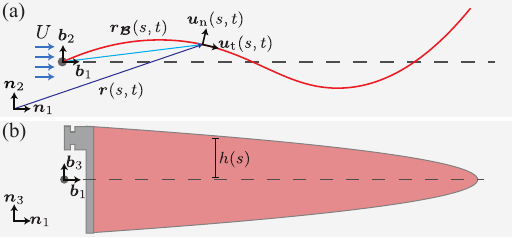}
\end{center}
\vspace{-2ex}    
\caption{\textbf{Kinematic definitions and geometry of the tail planform used for analysis and model formulation.}~\textbf{(a)}\,Idealized kinematics of the tail undulation during an actuation cycle. The inertial and \mbox{body-fixed} frames of reference are defined as \mbox{$\bs{\mathcal{N}} = \{\bs{n}_1,\bs{n}_2, \bs{n}_3\}$} and \mbox{$\bs{\mathcal{B}} = \{\bs{b}_1,\bs{b}_2,\bs{b}_3\}$}, respectively. To describe the motion of the tail as a traveling wave in the \mbox{$\bs{b}_1$--$\bs{b}_2$} plane, we define the Lagrangian coordinate, $s$, and the centerline relative to $\bs{\mathcal{N}}$ as \mbox{$\bs{r}(s,t) = \left[x(s,t)\,\,y(s,t)\,\,0 \right]^T$}, where \mbox{$t \geq 0$} denotes time. According to this model, the traveling wave \textit{travels} from left to right and the undulating tail travels at a constant speed $U$ from right to left; therefore, the centerline relative to $\bs{\mathcal{B}}$ can be described as \mbox{$\bs{r}_{\bs{\mathcal{B}}}(s,t) = \left[x(s,t)-Ut\,\,y(s,t)\,\,0 \right]^T$}. To track the local forces acting on the tail, we define the Lagrangian frame \mbox{$\left\{ \bs{u}_{\ts{t}},\bs{u}_{\ts{n}} \right\}$}, where $\bs{u}_{\ts{t}}(s,t)$ and $\bs{u}_{\ts{n}}(s,t)$ are the unit vectors tangent and normal to the tail's centerline.~\textbf{(b)}\,The planform of the undulating tails used to drive the propulsors considered in this paper have a parabolic shape with a local height given by \mbox{$h(s) = 0.694\sqrt{l-s}$}.\label{FIG02}}
\vspace{-2.0ex}
\end{figure}

\section{Reactive-Force Model} 
\vspace{-0.5ex}
\label{SECTION02}
In this section, we present the theoretical framework used to model the forces produced by \mbox{VLEIBot-type} propulsors. Employing the \mbox{large-amplitude} \mbox{elongated-body} approach presented in\cite{LighthillMJ1971}, we model the lateral undulation of a propulsive \textit{thin} tail as a traveling wave with the characteristics shown~in \mbox{Fig.\,\ref{FIG02}(a)}. As seen in \mbox{Figs.\,\ref{FIG02}(a)~and~(b)}, the triplet of unit vectors \mbox{$\bs{\mathcal{N}}=\{\bs{n}_1,\bs{n}_2,\bs{n}_3\}$} denotes the inertial frame of reference, fixed to the planet Earth, used to describe the kinematics of the tail and analyzing the dynamics of the fluid in its close proximity. Similarly, the triplet of unit vectors \mbox{$\bs{\mathcal{B}}=\{ \bs{b}_1,\bs{b}_2,\bs{b}_3\}$} denotes the \mbox{body-fixed} frame of reference. In agreement with these definitions, the directions of $\bs{b}_1$, $\bs{b}_2$, and $\bs{b}_3$ coincide with those of $\bs{n}_1$, $\bs{n}_2$, and $\bs{n}_3$ at any given instant, and $\bs{b}_1$ is collinear with the midline of the tail when in the extended condition---i.e., the planform of the tail is perpendicular to the \mbox{$\bs{n}_1$--$\bs{n}_2$} plane. Consistent with these idealizations, we assume that the tail does not twist during undulation and, therefore, that the forces acting along the $\bs{b}_3$ direction are negligible. To track the forces acting on the tail---all assumed inside the \mbox{$\bs{n}_1$--$\bs{n}_2$} plane---we define the Lagrangian coordinate \mbox{$s \in \left[0,l\right]$}, according to which the arc length $s$ is the distance from the origin of $\bs{\mathcal{B}}$---along the orthogonal projection of the tail onto \mbox{$\bs{n}_1$--$\bs{n}_2$}---to a vertical infinitesimal element of volume, and $l$ is the total length of the tail. 

The position of a point on the centerline of the tail, relative to the origin of $\bs{\mathcal{N}}$, at the Lagrangian coordinate $s$ and time $t$ is given by \mbox{$\bs{r}(s,t) = \left[x(s,t)~y(s,t)~0\right]^T$}. For modeling purposes, we assume that the tail moves from right to left at a constant speed $U$---information embedded in $x(s,t)$---and that the associated traveling wave travels from left to right, as shown in \mbox{Fig.\,\ref{FIG02}(a)}. Under these assumptions, $\bs{\mathcal{B}}$ is also an inertial frame. The tail material (AirTech~A$4000$R$14417$) is assumed inextensible and, as a consequence, \mbox{$\left\|\partial{\bs{r}}/\partial{s}\right\|_2 = 1$} and \mbox{$\left[ds/dt \right]_{\ts{tail}} = 0$}, for all \mbox{$(s,t)$}. Thus, the velocity of a point on the centerline of the tail can be found as
\begin{align}
\bs{v}_{\ts{b}}(s,t) = \frac{D \bs{r}}{Dt} = \frac{\partial \bs{r}}{\partial t} +  \frac{\partial \bs{r}}{\partial s} \left[ \frac{ds}{dt} \right]_{\ts{tail}} = \frac{\partial \bs{r}}{\partial t}.
\label{EQ01}
\end{align}
Also, we define a local Lagrangian frame, $\bs{\mathcal{L}}$, with its primary axis aligned with the unit tangent vector of the tail centerline, \mbox{$\bs{u}_{\ts{t}}(s,t)$}, and its secondary axis aligned with the unit normal vector of the tail centerline, $\bs{u}_{\ts{n}}(s,t)$, as shown in~\mbox{Fig.\,\ref{FIG02}(a)}. Using these definitions, (\ref{EQ01}) can be rewritten as
\begin{align}
\bs{v}_{\ts{b}}(s,t) = v_{\ts{b},\ts{n}}(s,t) \bs{u}_{\ts{n}}(s,t) + v_{\ts{b},\ts{t}}(s,t) \bs{u}_{\ts{t}}(s,t).
\label{EQ02}
\end{align}
Next, following Lighthill's approach, we write the velocity of a parcel of fluid at $(s,t)$ in $\bs{\mathcal{L}}$ coordinates as
\begin{align}
\bs{v}(s,t) = v_{\ts{n}}(s,t) \bs{u}_{\ts{n}}(s,t) + v_{\ts{t}}(s,t) \bs{u}_{\ts{t}}(s,t),
\label{EQ03}
\end{align}
where, \mbox{$v_{\ts{n}} = \bs{v} \cdot \bs{u}_{\ts{n}}$} and \mbox{$v_{\ts{t}} = \bs{v} \cdot \bs{u}_{\ts{t}}$}.

In this approach, we assume that the fluid is inviscid and that cannot penetrate the surface of the tail. From the \mbox{no-penetration} condition, it immediately follows that \mbox{$v_{\ts{n}} = v_{\ts{b},\ts{n}}$}; therefore, both $v_{\ts{n}}$ and ${\partial v_{\ts{n}}}/{\partial t} $ can be estimated from the kinematics of the tail's centerline. From the \mbox{inviscid-fluid} condition, it follows that, in general, \mbox{$v_{\ts{t}} \neq v_{\ts{b},\ts{t}}$}, which implies that the \mbox{no-slip} condition does not apply as a parcel of fluid can slide along the surface of the tail; therefore, \mbox{$v_{\ts{t}} = \left[ds/dt\right]_{\ts{fluid}} $} cannot be directly estimated from the kinematics of the tail's centerline.  

Reactive theory as presented \mbox{in\cite{LighthillMJ1971}} is based on the notion of \textit{added mass}, which is an idealization of the volume per unit length of incompressible fluid displaced when the modeled tail undulates while submerged in the surrounding medium. Here, we describe the added mass per unit length along the tail as  
\begin{align}
m(s) = 2 \rho \beta h(s),
\label{EQ04}
\end{align}
where $\rho$ is the density of the fluid; $\beta$ is the thickness of the tail material; and, \mbox{$h(s) = 0.694 \sqrt{l-s}$} defines the shape profile of the parabolic tail considered in this paper and shown in~\mbox{Fig.\,\ref{FIG02}(b)}. In this approach, \textit{resistive} forces associated with fluid flow produced by viscosity are assumed negligible. Thus, as explained in the Appendix, we only consider the forces acting on the tail that arise from the reaction to accelerating the corresponding added mass. Namely, using Newton's laws and including Lighthill's assumptions, we obtain that the total instantaneous reactive force exerted on the undulating tail is given by
\begin{align}
\bs{F}_{\ts{r}}(t) = -\int_0^l m(s) \left( \frac{\partial v_{\ts{n}}}{\partial t} +  v_{\ts{t}} \frac{\partial v_{\ts{n}}}{\partial s} \right) \bs{u}_{\ts{n}}(s,t) ds.
\label{EQ05}
\end{align}
Then, by extracting the component along the direction of locomotion, $\bs{b}_1$, we obtain the magnitude of the \textit{signed} instantaneous thrust force,
\begin{align}
F_{\ts{th}} = -\int_0^l m(s) \left( \frac{\partial v_{\ts{n}}}{\partial t} +  v_{\ts{t}} \frac{\partial v_{\ts{n}}}{\partial s} \right) \bs{u}_{\ts{n}} \cdot \bs{b}_1 ds,
\label{EQ06}
\end{align}
and its corresponding time average,
\begin{align}
\bar{F}_{\ts{th}} = -\int_0^l m(s) \left \langle  \left( \frac{\partial v_{\ts{n}}}{\partial t} +  v_{\ts{t}} \frac{\partial v_{\ts{n}}}{\partial s} \right)  u_{\ts{n},1} \right \rangle ds,
\label{EQ07}
\end{align}
where \mbox{$u_{\ts{n},1} = \bs{u}_{\ts{n}} \cdot \bs{b}_1$} and $\langle \cdot \rangle$ denotes the time average of the corresponding function over an undulation cycle. The integrals specified by (\ref{EQ06}) and (\ref{EQ07}) can be separated into an unsteady term and a convective term. The unsteady term represents the reaction force exerted on the tail due to direct normal acceleration of the fluid parcel in contact with it; this term can be significant even when the undulating tail does not travel relative to the inertial frame---\mbox{i.e.,~$U=0$}---and the instantaneous curvature of the tail is small. In contrast, the convective term can be considered negligible for \mbox{$U=0$} and a small curvature; additionally, it is difficult to estimate $v_{\ts{t}}$ from video footage without the use of advanced techniques such as \textit{particle image velocimetry} (PIV). Therefore, for the experiments and numerical estimations presented in~\mbox{Section\,\ref{SECTION04}}, we consider that \mbox{$v_{\ts{t}}=0$} because thrust is measured with the undulating tail attached to a static sensor. 

As discussed in\cite{LighthillMJ1971}, over an undulation cycle---through a traveling wave like the one depicted in \mbox{Fig.\,\ref{FIG02}(a)}---the actuator delivers a finite amount of mechanical energy at the connection with the tail. Therefore, assuming that the elastic energy stored in the tail material is negligible, the \mbox{cycle-averaged} power carried by a traveling wave along the tail can be estimated as 
\begin{align}
P_{\ts{w}}(s) = \frac{1}{2}m(s)\left\langle v_{\ts{n}}^2(s,t) \right\rangle v_{\ts{w}}, \label{EQ08}
\end{align}
where $v_{\ts{w}}$ is the traveling velocity of the wave, which is determined by the actuation frequency. Using basic kinematics, we conclude that to maximize the stroke amplitude during operation, \mbox{$\left\langle v_{\ts{n}}^2(s,t) \right\rangle$} must increase and, due to energy dissipation, $P_{\ts{w}}(s)$ is expected to decrease along $s$. This observation indicates that a \mbox{high-performance} propulsion system of the type considered here must necessarily combine \textit{high-work-density}\,(HWD) actuation and a \mbox{soft-tail} design with a decreasing $h(s)$ along $s$. As already mentioned, the model specified by (\ref{EQ06}) and (\ref{EQ07}) neglects resistive forces such as \mbox{skin-friction} and \mbox{boundary-layer} dissipation, which become more dominant as the Reynolds number decreases. Therefore, to describe force generation at low Reynolds numbers (\mbox{$\ts{Re} < 1$}), \mbox{resistive-force} models are necessary. The propulsors studied in the presented research operate at Reynolds numbers in the range from about $300$ to $900$, and it is not immediately obvious that the approach presented here is suitable for this flow regime. To assess this and other issues, we developed the \mbox{{\textmu}N-resolution} force sensor discussed in the next section. 
\begin{figure}[t!]
\vspace{1.4ex}
\begin{center}
\includegraphics[width=0.48\textwidth]{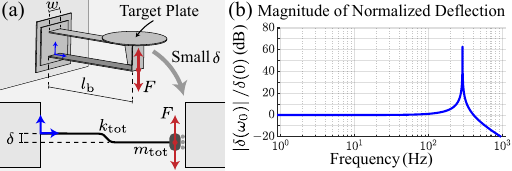}
\end{center}
\vspace{-2ex}
\caption{\hspace{-2ex}\textbf{Idealization and dynamic modeling of the DCS constituting the \mbox{{\textmu}N-resolution} force sensor used to measure thrust.} \textbf{(a)}\,Simplified dynamic model of the DCS made of \mbox{Invar-$36$}.~\textbf{(b)}\,Magnitude of the normalized sensing structure's response, \mbox{$\left| \delta(\omega_0)\right|/\delta(0)$} in decibels (dB), to a periodic excitation of the form \mbox{$F(t) = F_0 e^{j\omega_0t}$}, for \mbox{$j = \sqrt{-1}$}. \label{FIG03}}
\vspace{2.0ex}
\begin{center}
\includegraphics[width=0.48\textwidth]{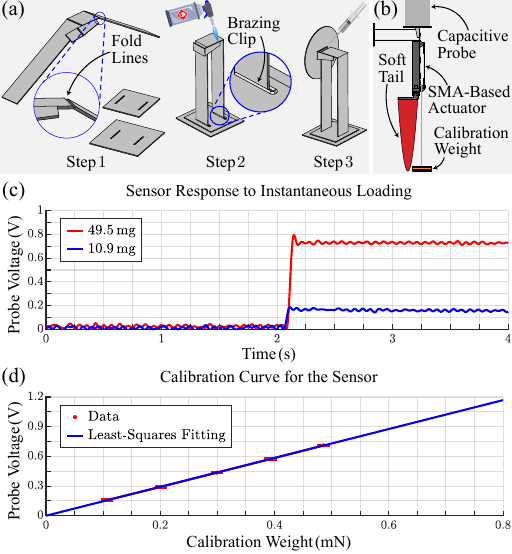}
\end{center}    
\vspace{-3ex}
\caption{\textbf{Fabrication and calibration of the force sensor.}~\textbf{(a)}\,The sensor is fabricated in three steps. In \mbox{Step\,$1$}, the precursor pieces of the sensor---made of \mbox{$0.153$-mm-thick} \mbox{Invar-$36$ material}---are premachined and folded along perforated lines to create $3$D structural components with assembly features; in \mbox{Step\,$2$}, clips made of \mbox{$80/16/4$\,wt.\%} \mbox{Ag-Cu-Zn} brazing wire are placed at the interfaces between pieces and folding lines to weld the structure together, according to the method introduced in\cite{Singer2019ClipBrazing}; last, in \mbox{Step\,$3$}, a \mbox{$1$-cm-diameter} circular target plate---required to operate the capacitive displacement sensor used to measure the deflection of the DCS depicted in \mbox{Fig.\,\ref{FIG03}(a)}---is mirror polished and connected to the deflecting structure, using conductive silver epoxy (MG\,Chemicals\,\mbox{$8331$D}).~\textbf{(b)}\,To calibrate the sensor, we hung sequentially a series of \mbox{custom-made} laboratory weights, using a linear stage, to generate a set of \mbox{Heaviside-like} inputs. The resulting step responses are used to find a static mapping between weight and the voltage outputted by the capacitive sensor.~\textbf{(c)}\,Voltages outputted by the capacitance displacement sensor as responses to step loadings of \mbox{$10.9$\,mg} (\mbox{$0.107$\,mN}) and \mbox{$49.5$\,mg}~(\mbox{$0.393$\,mN}).~\textbf{(d)}\,Resulting calibration curve. In this plot, each data point indicates the mean and ESD of the values of ten \mbox{back-to-back} \mbox{steady-state} step responses obtained using a weight in the set \mbox{$\left\{0.107,0.2011,0.299,0.393,0.486\right\}$\,mN}. The blue line represents the \mbox{least-squares} fitting of the data. The resulting static mapping was found to be \mbox{$1.46$\,V$\cdot$\,mN$^{-1}$} with a coefficient of determination of $0.999$. 
\label{FIG04}}
\vspace{-2.0ex}
\end{figure}

\section{A New \mbox{\unit{\micro\text{N}}-Resolution} Force Sensor}
\vspace{-0.5ex}
\label{SECTION03}
\subsection{Dynamic Modeling for Design} 
\vspace{-0.5ex}
\label{SECTION02A}
From simple experiments, we estimate that the instantaneous hydrodynamic forces generated by the microswimmers shown in \mbox{Fig.\,\ref{FIG01}} must exhibit oscillatory components with sizes as small as \mbox{$6$\,\textmu{N}} over the range of $0$ to \mbox{$1.0$\,mN}. Therefore, the sensor we presented in\cite{Singer2019ClipBrazing}---with a resolution of \mbox{$7.48$\,\textmu{N}}---is not well suited for the task at hand. This limitation prompted us to improve the resolution of existing \mbox{{\textmu}N-resolution} force sensors for microrobotic applications. The resulting sensor presented here has a resolution of \mbox{$0.274$\,{\textmu}N} and an operational range of $-6.85$ to \mbox{$6.85$\,mN}. To develop this new device, we one more time followed the analysis and design principles for \textit{\mbox{dual-cantilever} structures} (DCSs) discussed in\cite{Singer2019ClipBrazing,Hoffman2009MuliAxis}, and references therein. A lumped model for a structure of this type is depicted in \mbox{Fig.\,\ref{FIG03}(a)}, in which the target plate is necessary to measure the deformation of the distal end of the DCS using a capacitive sensor and, by convention, the upward and downward directions are considered to be positive and negative, respectively. According to the idealization shown in \mbox{Fig.\,\ref{FIG03}(a)}, a static mapping between the force value, $F$---defined as the signed magnitude of the corresponding vector force---and the displacement of the structure's target plate, $\delta$, follows Hooke's law, \mbox{$\delta = F/k_{\ts{tot}}$}, where the elastic constant can be calculated as \mbox{$k_{\ts{tot}} = 24EI/l_{\ts{b}}^3$}. In this description, $E$ is the Young modulus of the structure's material (\mbox{Invar-$36$}); $l_{\ts{b}}$ is the length of the DCS, as defined in \mbox{Fig.\,\ref{FIG03}(a)}; and, $I$ is the moment of inertia of each composing beam's \mbox{cross-sectional} area given by \mbox{$I = wd^3/12$}, where $w$ is the width of the beam, as defined in Fig.\,\ref{FIG03}(a), and $d$ is the thickness of the Invar-$36$ material. Using basic beam theory, the DCS can be modeled as an equivalent \mbox{mass-spring-damper} system, whose response to a periodic excitation of the form \mbox{$F(t) = F_0 e^{j\omega_0t}$}, with \mbox{$j = \sqrt{-1}$}, has a magnitude given by
\begin{align}
\left|\delta(\omega_0) \right| = \frac{F_0}{k_{\ts{tot}}\left[\eta^2 + (1-\omega_0^2/\omega_{\ts{n}}^2)^2\right]^{\frac{1}{2}}}, 
\label{EQ09}
\end{align}
in which $F_0$ is a known constant amplitude; $\omega_0$ is a known constant frequency; \mbox{$\eta = 0.007$} is a hysteretic loss coefficient used to account for energy losses, provided by the manufacturer of the \mbox{Invar-$36$} material; and, \mbox{$\omega_{\ts{n}}=\sqrt{k_{\ts{tot}}/m_{\ts{tot}}}$} is the undamped natural frequency of the equivalent dynamics of the structure, with the equivalent mass $m_{\ts{tot}}$. During development, through an iterative design process, we kept the sensor dimensions reported in\cite{Singer2019ClipBrazing} unchanged---i.e., \mbox{$w = 3$\,mm}, \mbox{$d=0.153$\,mm}---and varied the length, $l_{\ts{b}}$, which we finally selected to be \mbox{$16$\,mm}. The normalized magnitude response, \mbox{$\left| \delta(\omega_0)\right|/\delta(0)$}, in decibels (dB) for the resulting design across the frequency set \mbox{$\left\{ \omega_0/2\pi \in \left[10^0, 10^3\right]\right\}$\,Hz} is shown in~\mbox{Fig.\,\ref{FIG03}(b)}. As seen, the natural frequency of the system is \mbox{$288$\,Hz}, which is sufficiently high to obtain a wide useful measurement band for the desired application and flexibility to achieve a high resolution, precision, and accuracy. In this case, we predict a measurement error of less than \mbox{$1$\hspace{0.1ex}\%} for excitation frequencies below \mbox{$28.7$\,Hz}.

\subsection{Fabrication of the Sensor} 
\vspace{-0.5ex}
\label{SECTION02B}
The fabrication process of the sensor, which consists of three steps, is graphically explained in \mbox{Fig.\,\ref{FIG04}(a)}. In \mbox{Step\,$1$}, the \mbox{Invar-$36$} pieces of the DCS are premachined using a \mbox{$3$-W} UV laser (Photonics Industries \mbox{DCH-$355$-$3$}), and folded along perforated lines to create \mbox{$3$D} structures; in \mbox{Step\,$2$}, following the \mbox{clip-brazing} method introduced in\cite{Singer2019ClipBrazing}, small clips of brazing wire---with a \mbox{$80/16/4$\,wt.\%} \mbox{Ag-Cu-Zn} composition---are placed at the interfaces between pieces and folding lines, and then melted with a butane torch, to create a stable rigid monolithic structure; lastly, in \mbox{Step\,$3$}, the \mbox{$1$-cm-diameter} target plate, also made of \mbox{Invar-$36$}, depicted in \mbox{Fig.\,\ref{FIG03}(a)} is mirror polished and connected to the DCS as depicted in \mbox{Fig.\,\ref{FIG03}(a)}, using conductive silver epoxy (MG~Chemicals~$8331$D). After fabrication, to prevent oxidation, we mechanically polished and coated the structure with oil (WD-$40$). As seen in \mbox{Fig.\,\ref{FIG04}(b)}, before performing a force characterization experiment, the tested propulsor is glued to the lower distal end of the DCS, and the target plate is precisely aligned beneath the cylindrical probe of the displacement capacitive sensor (Physik Instrumente \mbox{D-$510.021$}) employed to measure the instantaneous value of $\delta$.  
\begin{figure}[t!]
\vspace{1.5ex}
\begin{center}    
\includegraphics[width=0.48\textwidth]{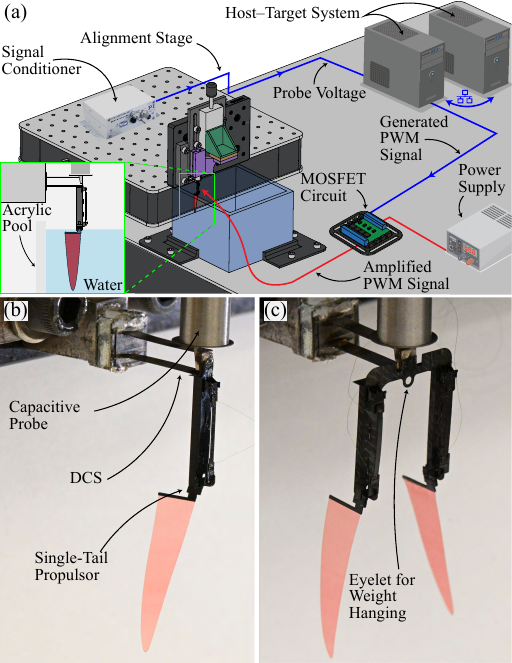}
\end{center}
\vspace{-2ex}
\caption{\textbf{Experimental setup used for characterization of the force generated by a propulsor during operation.}~\textbf{(a)}\,A Mathworks Simulink \mbox{Real-Time} \mbox{host--target} system, equipped with a National Instruments \mbox{PCI-$6229$} \mbox{AD/DA} board, is used to generate, process, and record signals at a rate of \mbox{$5$\,kHz}. The PWM signal required for actuation is generated using the \mbox{AD/DA} board of the \mbox{host--target} system; then, this signal is \mbox{power-amplified} with a \mbox{MOSFET-based} circuit that acts as a switch that opens and closes the electrical paths to the actuator from a power supply. After the tested propulsor is attached to the sensing structure, as depicted in the inset, the target plate (see \mbox{Fig.\,\ref{FIG03}(a)}) is aligned beneath the probe of the capacitive sensor (Physik\,Instrumente\,D-$519.021$) that measures the deflection of the DCS. Then, the voltage outputted by the capacitive sensor is filtered through a signal conditioner (Physik Instrumente E-$852$ PISeca) and recorded using the AD/DA board of the \mbox{host--target} system. During testing, the propulsor's tail remains submerged from its leading edge downward in an acrylic pool filled with water, as depicted in the inset.~\textbf{(b)}\,Photograph of the tested \mbox{single-tail} propulsor attached to the DCS aligned beneath the probe of the capacitive displacement sensor.~\textbf{(c)}\,Photograph of the tested \mbox{dual-tail} propulsor attached to the DCS aligned beneath the probe of the capacitive displacement sensor. \label{FIG05}}
\vspace{-2ex}
\end{figure}
\begin{figure*}[t!]
\vspace{1.5ex}
\begin{center}
\includegraphics[width=0.98\textwidth]{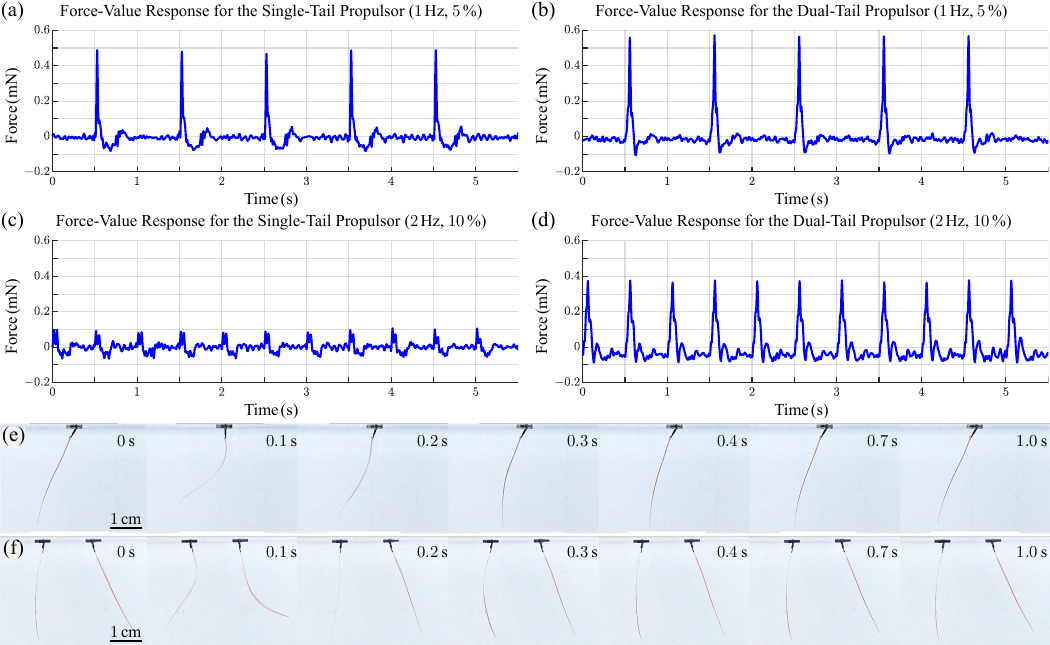}
\end{center}    
\vspace{-2ex}
\caption{\hspace{-1ex}\textbf{Data and photographic sequences corresponding to two force characterization experiments.}~\textbf{(a)}\,Measured instantaneous force generated by the tested \mbox{single-tail} propulsor over \mbox{$5.5$\,s} of \mbox{steady-state} operation, using a \mbox{$1$-Hz} PWM actuation signal with a DC of \mbox{$5$\hspace{0.1ex}\%}. In this case, the means of the peak and \mbox{cycle-averaged} forces, $\bar{F}_{\ts{p}}$ and $\bar{F}_{\ts{a}}$, are \mbox{$0.48$\,mN} and \mbox{$3.6$\,{\textmu}N}, respectively.~\textbf{(b)}\,Measured instantaneous force generated by the tested \mbox{dual-tail} propulsor over \mbox{$5.5$\,s} of \mbox{steady-state} operation, using a \mbox{$1$-Hz} PWM actuation signal with a DC of \mbox{$5$\hspace{0.1ex}\%}. In this case, the means of the peak and \mbox{cycle-averaged} forces, $\bar{F}_{\ts{p}}$ and $\bar{F}_{\ts{a}}$, are \mbox{$0.58$\,mN} and \mbox{$12.6$\,{\textmu}N}, respectively.~\textbf{(c)}\,Measured instantaneous force generated by the tested \mbox{single-tail} propulsor over \mbox{$5.5$\,s} of \mbox{steady-state} operation, using a \mbox{$2$-Hz} PWM actuation signal with a DC of \mbox{$10$\hspace{0.1ex}\%}. In this case, the means of the peak and \mbox{cycle-averaged} forces, $\bar{F}_{\ts{p}}$ and $\bar{F}_{\ts{a}}$, are \mbox{$0.084$\,mN} and \mbox{$-0.72$\,{\textmu}N}, respectively.~\textbf{(d)}\,Measured instantaneous force generated by the tested \mbox{dual-tail} propulsor over \mbox{$5.5$\,s} of \mbox{steady-state} operation, using a \mbox{$2$-Hz} PWM actuation signal with a DC of \mbox{$10$\hspace{0.1ex}\%}. In this case, the means of the peak and \mbox{cycle-averaged} forces, $\bar{F}_{\ts{p}}$ and $\bar{F}_{\ts{a}}$, are \mbox{$0.39$\,mN} and \mbox{$11.4$\,{\textmu}N}, respectively.~\textbf{(e)}\,Sequence of video frames---taken at intervals of \mbox{$0.1$\,s}---of the full actuation cycle corresponding to the first second of the experimental data presented in~(a).~\textbf{(f)}\,Sequence of video frames---taken at intervals of \mbox{$0.1$\,s}---of the full actuation cycle corresponding to the first second of the experimental data presented in~(b). Video footage of these experiments can be viewed in the accompanying supplementary movie. \label{FIG06}}
\vspace{-0.2ex}
\end{figure*}

\subsection{Calibration of the Sensor}
\vspace{-0.5ex}
\label{SECTION3C}
Before operation, we calibrated the sensor using the same procedure described in \cite{Singer2019ClipBrazing}. Under this methodology, as depicted in \mbox{Fig.\,\ref{FIG04}(b)}, a series of \mbox{custom-made} laboratory weights are sequentially hung from the DCS with a tested propulsor attached to it in order to find a relationship between weight and the voltage outputted by the capacitive displacement sensor. To mitigate the chances of human error, during each calibration test, we hang the corresponding laboratory weight almost instantaneously using a linear stage. This action smoothly excites the \mbox{second-order} dynamics of the DCS with a \mbox{Heaviside-like} force, thus generating a step response in the sensor reading. Two examples of such responses are shown in \mbox{Fig.\,\ref{FIG04}(c)}; here, the signals in blue and red correspond to calibration weights with masses of \mbox{$10.9$\,mg} and \mbox{$49.5$\,mg}, respectively. The resulting calibration mapping used to interpret the data obtained through the experiments presented in \mbox{Section\,\ref{SECTION04}} is shown in \mbox{Fig.\,\ref{FIG04}(d)}. In this plot, each red data point indicates the mean and \textit{experimental standard deviation}~(ESD) of the \mbox{steady-state} values corresponding to the ten step responses obtained through ten \mbox{back-to-back} experiments performed for each calibration weight in the set \mbox{$\left\{0.107,0.2011,0.299,0.393,0.486\right\}$\,mN}. Also, in \mbox{Fig.\,\ref{FIG04}(d)}, the blue line represents the \mbox{least-squares} fitting of the data. This linear function maps the probe voltage generated by the capacitive displacement sensor, $V_{\ts{p}}$, to the sensed force, $F_{\ts{s}}$, according to \mbox{$V_{\ts{p}} = 1.46\cdot F_{\ts{s}}$}, with a coefficient of determination of $0.999$. We performed the calibration procedure several times with both the \mbox{single-tail} and \mbox{dual-tail} propulsors connected to the sensor as depicted in \mbox{Fig.\,\ref{FIG04}(b)}, and we obtained almost identical results. For this reason, we used the same mapping in \mbox{Fig.\,\ref{FIG04}(b)} to interpret all the force data discussed in Section\,\ref{SECTION04}.

\section{Experiments for Force Characterization}
\vspace{-0.5ex}
\label{SECTION04}
\subsection{Experimental Setup and Data Processing} 
\vspace{-0.5ex}
\label{SECTION04A}
The experimental setup used for characterizing the forces produced by both the \mbox{single-tail} and \mbox{dual-tail} tested propulsors is shown in \mbox{Fig.\,\ref{FIG05}(a)}. Here, a Mathworks Simulink \mbox{Real-Time} \mbox{host--target} system, equipped with an \mbox{\textit{analog-digital/digital-analog}} (AD/DA) board (National Instruments PCI-$6229$), is used to generate, process, and record signals at a rate of \mbox{$5$\,kHz}. During operation, the target computer generates the \textit{pulse-width modulation} (PWM) signal required for actuation, with prescribed \textit{frequency} ($f$) and \textit{duty-cycle} (DC) parameters. This signal is \mbox{power-amplified} by a \mbox{MOSFET-based} circuit (YYNMOS-$4$), which opens and closes the conductive pathways from a power supply to the actuation mechanism of the tested propulsor. To obtain the desired amplitudes of actuation, the voltage of the power supply is adjusted according to the procedure discussed in\cite{TrygstadCK2024}. As seen in the inset of \mbox{Fig.\,\ref{FIG05}(a)}, during a force characterization experiment, the tested propulsor remains submerged from just above the leading edge of its soft tail in a pool filled with water. During the experiments, we tested both studied propulsors at all the combined \mbox{$f$--DC} pairs in the sets \mbox{$f\in\{1:1:4\}$\,Hz} and \mbox{$\ts{DC}\in\{1:10\}$\,\%}. To ensure similar operational tail strokes across different experiments and devices, we tuned the \mbox{on-height} voltages of the exciting PWM signals such that the current flowing through the SMA wires was \mbox{$250$\,mA} in all the cases. The photographs in \mbox{Figs.\,\ref{FIG05}(b)~and~(c)} show the \mbox{single-tail} and \mbox{dual-tail} tested propulsors connected to the force sensor, before operation. 
\begin{figure*}
\vspace{1.5ex}
\begin{center}
\includegraphics[width=0.98\textwidth]{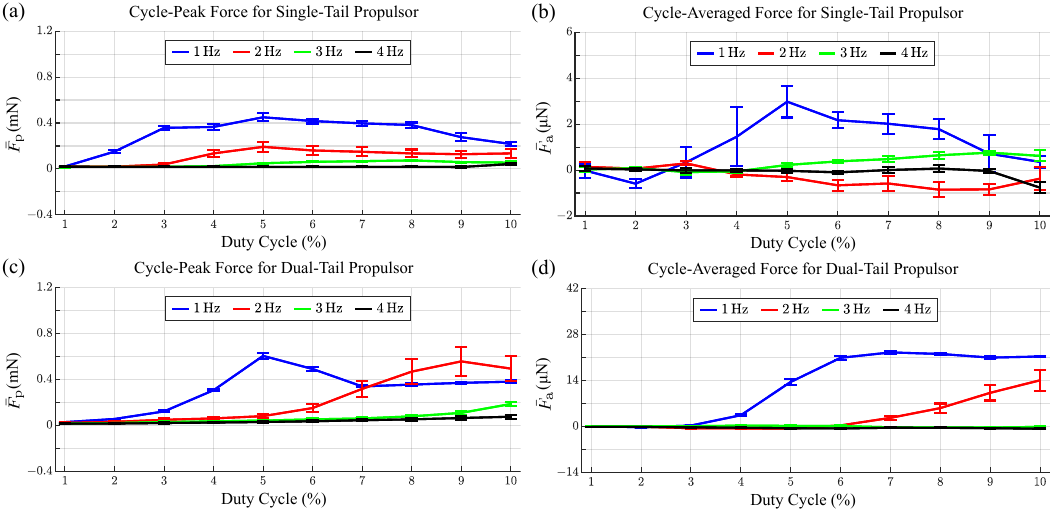}
\end{center}
\vspace{-3ex}
\caption{\textbf{Force data obtained through characterization experiments.}~\textbf{(a)}~In this plot, each data point indicates the mean and ESD of the $\bar{F}_{\text{p}}$ values corresponding to five \mbox{back-to-back} tests, performed using the \mbox{single-tail} propulsor and a \mbox{$f$--DC} pair specified by color and position.~\textbf{(b)}~In this plot, each data point indicates the mean and ESD of the $\bar{F}_{\text{a}}$ values corresponding to five \mbox{back-to-back} tests, performed using the \mbox{single-tail} propulsor and a \mbox{$f$--DC} pair specified by color and position.~\textbf{(c)}~In this plot, each data point indicates the mean and ESD of the $\bar{F}_{\text{p}}$ values corresponding to five \mbox{back-to-back} tests, performed using the \mbox{dual-tail} propulsor and a \mbox{$f$--DC} pair specified by color and position.~\textbf{(d)}~In this plot, each data point indicates the mean and ESD of the $\bar{F}_{\text{a}}$ values corresponding to five \mbox{back-to-back} tests, performed using the \mbox{dual-tail} propulsor and a \mbox{$f$--DC} pair specified by color and position. \label{FIG07}}
\vspace{-2.0ex}
\end{figure*}

During operation, the voltages outputted by the capacitive displacement sensor are filtered through a signal conditioner (Physik Instrumente \mbox{E-$852$} PISeca) before being sampled by the \mbox{AD/DA} board of the \mbox{host--target} system and recorded. For the data presented in this section, after collection, the signals were digitally processed to remove \mbox{high-frequency} electrical noise as well as \mbox{low-frequency} sensor drifting possibly caused by changes in the environmental temperature and humidity. More specifically, the raw data was filtered through a \mbox{low-pass} \textit{\mbox{finite-impulse} response}~(FIR) \mbox{zero-phase} digital filter of order \mbox{$5 \times 10^3$} with a cutoff frequency of \mbox{$0.01$\,Hz} to extract the \mbox{low-frequency} drift. This extracted drift was then subtracted from the raw signal to obtain a \mbox{drift-free} signal. This \mbox{drift-free} signal was then filtered through another \mbox{low-pass} FIR \mbox{zero-phase} digital filter of order \mbox{$2\times10^3$} with a cutoff frequency of \mbox{$50$\,Hz} to cancel \mbox{high-frequency} noise, which yields the final filtered data used for analysis. We define the \mbox{cycle-peak} force, $F_{\ts{p}}$, as the maximum force value produced by a propulsor during an actuation cycle, and the \mbox{cycle-averaged} force, $F_{\ts{a}}$, as the average of the force values measured over the same actuation cycle. For each independent force test, we used five \mbox{steady-state} actuation cycles to compute the mean of the corresponding five $F_{\ts{p}}$ values, which we denote as $\bar{F}_{\ts{p}}$; similarly, for each independent force test, we used five \mbox{steady-state} actuation cycles to compute the mean of the corresponding five $F_{\ts{a}}$ values, which we denote as $\bar{F}_{\ts{a}}$. For the purposes of data analysis, for each \mbox{$f$--DC} pair, we performed five independent force tests and we computed the means and ESDs of the $\bar{F}_{\ts{p}}$ and $\bar{F}_{\ts{a}}$ values.

\subsection{Results and Discussion} 
\vspace{-0.5ex}
\label{SECTION04B}
\mbox{Figs.\,\ref{FIG06}(a)~and~(b)} show the measured instantaneous forces respectively generated by the \mbox{single-tail} and \mbox{dual-tail} tested propulsors, using a \mbox{$1$-Hz} PWM actuation signal with a DC of \mbox{$5$\hspace{0.1ex}\%}. In these cases, it can be observed that the \mbox{dual-tail} propulsor generates noticeably higher peaks of positive force, while the \mbox{single-tail} propulsor generates deeper valleys of negative force. These two phenomena combined indicate that the average force in the \mbox{single-tail} case must be lower than in the \mbox{dual-tail} case. \mbox{Figs.\,\ref{FIG06}(c)~and~(d)} show the instantaneous forces measured for the \mbox{single-tail} and \mbox{dual-tail} tested propulsors, using a \mbox{$2$-Hz} PWM actuation signal with a DC of \mbox{$10$\hspace{0.1ex}\%}. In these cases, it can be observed that the \mbox{dual-tail} propulsor generates drastically higher peaks of positive force, and both propulsors generate visually comparable valleys of negative force. These observations indicate that the average force in the \mbox{single-tail} case must be considerably lower than in the \mbox{dual-tail} case. \mbox{Figs.\,\ref{FIG06}(e)~and~(f)} show sequences of video frames, taken at \mbox{$0.1$-s} intervals, of full actuation cycles corresponding to the first second of experimental data presented in \mbox{Figs.\,\ref{FIG06}(a)~and~(b)}, respectively.

The data obtained through the force characterization experiments are summarized in \mbox{Fig.\,\ref{FIG07}}. Here, each data point in \mbox{Fig.\,\ref{FIG07}(a)} indicates the mean and ESD of the $\bar{F}_{\ts{p}}$ values corresponding to five \mbox{back-to-back} force tests performed using the \mbox{single-tail} propulsor and a given \mbox{$f$--DC} pair; and, each data point in \mbox{Fig.\,\ref{FIG07}(b)} indicates the mean and ESD of the $\bar{F}_{\ts{a}}$ values corresponding to the same five \mbox{back-to-back} force tests associated with the data shown in \mbox{Fig.\,\ref{FIG07}(a)}. Similarly, each data point in \mbox{Fig.\,\ref{FIG07}(c)} indicates the mean and ESD of the $\bar{F}_{\ts{p}}$ values corresponding to five \mbox{back-to-back} force tests performed using the \mbox{dual-tail} propulsor and a given \mbox{$f$--DC} pair; and, each data point in \mbox{Fig.\,\ref{FIG07}(d)} indicates the mean and ESD of the $\bar{F}_{\ts{a}}$ values corresponding to the same five \mbox{back-to-back} force tests associated with the data shown in \mbox{Fig.\,\ref{FIG07}(c)}. These data clearly show that, in all \mbox{single-tail} cases, the means of the $\bar{F}_{\ts{p}}$ values obtained with actuation frequencies of $3$ and \mbox{$4$\,Hz} are significantly lower than those measured at frequencies of $1$ and \mbox{$2$\,Hz}; while, in most but not all \mbox{single-tail} cases, the means of the $\bar{F}_{\ts{a}}$ values at \mbox{$1$\,Hz} are noticeably higher than those corresponding to the other three actuation frequencies. By comparison, in all the \mbox{dual-tail} cases, both the means of the $\bar{F}_{\ts{p}}$ and $\bar{F}_{\ts{a}}$ values obtained with actuation frequencies of $3$ and \mbox{$4$\,Hz} are significantly lower than those measured at frequencies of $1$ and \mbox{$2$\,Hz}. 

Furthermore, it can be observed that while the means of the $\bar{F}_{\ts{p}}$ values are generally comparable between the \mbox{single-tail} and \mbox{dual-tail} cases, the means of the $\bar{F}_{\ts{a}}$ values are significantly larger in the \mbox{dual-tail} case. As seen, among all \mbox{single-tail} cases, the maximum means of the $\bar{F}_{\ts{p}}$ and $\bar{F}_{\ts{a}}$ values both occur at a frequency of \mbox{$1$\,Hz} and DC of \mbox{$5$\hspace{0.1ex}\%}, which were measured to be \mbox{$0.45$\,mN} and \mbox{$2.97$\,{\textmu}{N}}, respectively. By comparison, among the \mbox{dual-tail} cases, the maximum mean of the $\bar{F}_{\ts{p}}$ values---\mbox{$0.61$\,mN}---occurs at an actuation frequency of \mbox{$1$\,Hz} and DC of \mbox{$5$\hspace{0.1ex}\%}, while the maximum mean of the $\bar{F}_{\ts{a}}$ values---\mbox{$22.6$\,{\textmu}N}---occurs at an actuation frequency of \mbox{$1$\,Hz} and DC of \mbox{$7$\hspace{0.1ex}\%}. Additionally, it is interesting to note that, at an actuation frequency of \mbox{$2$\,Hz}, the \mbox{dual-tail} propulsor performs drastically better than its \mbox{single-tail} counterpart; specifically, among all the \mbox{$2$-Hz} cases, the maximum mean of the $\bar{F}_{\ts{p}}$ values for the \mbox{dual-tail} propulsor---\mbox{$0.56$\,mN}, occurring at a DC of \mbox{$9$\hspace{0.1ex}\%}---is $4.31$ times larger than that corresponding to the \mbox{single-tail} at a DC of \mbox{$9$\hspace{0.1ex}\%}. 

Similarly, among all the \mbox{$2$-Hz} cases, the maximum mean of the $\bar{F}_{\ts{a}}$ values for the \mbox{dual-tail} propulsor---\mbox{$14.08$\,{\textmu}N}, occurring at a DC of \mbox{$10$\hspace{0.1ex}\%}---is positive and $37.01$ times larger than that corresponding to the \mbox{single-tail} at a DC of \mbox{$10$\hspace{0.1ex}\%}, which is, additionally, negative. The data presented in \mbox{Fig.\,\ref{FIG07}} provide compelling evidence demonstrating that the tested \mbox{dual-tail} propulsor produces considerably higher average thrusts than its \mbox{single-tail} counterpart, which explains to a significant extent the improved locomotion performance achieved by the \mbox{dual-tail} VLEIBot\textsuperscript{+} and VLEIBot\textsuperscript{++} swimmers compared to that measured for the \mbox{single-tail} VLEIBot swimmer. Note, however, that the presented force tests do not account for a periodic lateral motion experienced by the \mbox{single-tail} VLEIBot during swimming, a phenomenon likely produced by the periodic reactive torque acting on the robot's head as the tail flaps in the surrounding fluid. Furthermore, since the front rigid parts of the tested propulsors are held fixed relative to the inertial frame of reference, the forces measured according to the method graphically described in \mbox{Fig.\,\ref{FIG05}} are not fully representative of the total thrust produced by a swimmer moving forward at a constant speed \mbox{$U \neq 0$}. 
\begin{figure}
\vspace{1.5ex}
\begin{center}
\includegraphics[width=0.48\textwidth]{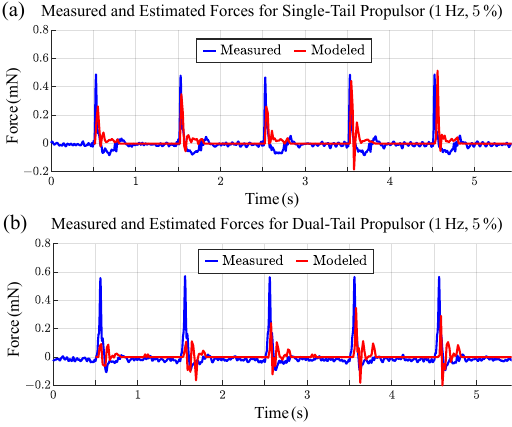}
\end{center}
\vspace{-3ex}    
\caption{\textbf{Comparison of the instantaneous \mbox{thrust-force} values obtained through direct measurement and \mbox{model-based} estimation.}~\textbf{(a)}\,Force values corresponding to the tested \mbox{single-tail} propulsor for a \mbox{$1$-Hz} PWM excitation signal with a DC of \mbox{$5$\hspace{0.1ex}\%}.~\textbf{(b)}\,Force values corresponding to the tested \mbox{dual-tail} propulsor for a \mbox{$1$-Hz} PWM excitation with a DC of \mbox{$5$\hspace{0.1ex}\%}. \label{FIG08}}
\vspace{-2.0ex}
\end{figure}

Last, we evaluate the predictive power and limitations of the model specified by (\ref{EQ06}) with \mbox{$v_{\ts{t}}=0$}, using a custom algorithm implemented in MATLAB that extracts the curvature of the tested tail from video footage. Specifically, this algorithm reads each video frame, converts it into a binary image, removes unwanted features, and then extracts the instantaneous centerline of the tested tail. The extracted instantaneous curvature of the tail is then used to estimate the kinematic variables required by the model specified by (\ref{EQ06}), assuming \mbox{$v_{\ts{t}}=0$}. \mbox{Figs.\,\ref{FIG08}(a)~and~(b)} compare the instantaneous thrust forces obtained through direct measurement and \mbox{model-based} estimation corresponding to a \mbox{$1$-Hz} PWM excitation with a DC of \mbox{$5$\hspace{0.1ex}\%}, for the \mbox{single-tail} and \mbox{dual-tail} propulsors, respectively. It can be clearly observed in \mbox{Fig.\,\ref{FIG08}(a)} that, for the \mbox{single-tail} case, the \mbox{model-based} estimated and measured forces are similar in shape and comparable in magnitude. However, the model predicts a slight positive force after the initial spike; in contrast, the measured signal exhibits a negative force valley during the same interval of the actuation cycle. In the \mbox{dual-tail} case, it can be observed that the amplitude of the \mbox{model-based} estimate is considerably lower than that measured experimentally. We hypothesize that this discrepancy was partly caused by the introduction of noise during the \mbox{image-processing-based} extraction of the tail's curvature from video footage. We believe, however, that the most important reason for the underestimation of the \mbox{model-based} thrust is that the operator specified by (\ref{EQ06}), with \mbox{$v_{\ts{t}}=0$}, neglects the convective contribution associated with the $v_{\ts{t}} {\partial u_{\ts{n}}}/{\partial s}$ term. To see the logic behind this argument, note that, in the \mbox{dual-tail} case, $v_{\ts{t}}$ is not expected to be negligible even though \mbox{$U=0$} because of interactions between the wake structures generated by the two undulating tails of the propulsors. It is well known that when wake structures interact, constructive or destructive patterns form, which might enhance or attenuate the process of thrust production\cite{TriantafyllouM1995}. Besides these limitations, the results presented in this paper are significant as they suggest that direct measurement of thrust through static experiments can be leveraged to design \mbox{single-tail} propulsors. Nonetheless, the experimental data and associated analyses highlight the need for further investigation into the modeling of the involved hydrodynamic forces to account for \mbox{multi-tail} \mbox{wake--structure} interaction and other convective phenomena. 

\section{Conclusion}
\vspace{-0.5ex}
\label{SECTION05}
We presented a study of the characteristics of the thrust generated by two newly developed \mbox{insect-scale} propulsors that leverage the FSI between unactuated soft tails and their surrounding fluid to function. The first actuator studied is \mbox{single-tailed}, and the second is double-tailed. In this approach, a driving actuator generates a wave that travels along a tail to produce hydrodynamic forces useful for locomotion. To describe the mechanism of thrust generation, we performed an analysis based on reactive theory that provides insights for the design and development of \mbox{high-performance} \mbox{insect-scale} propulsors. To experimentally measure the instantaneous thrust generated by the two studied devices during static tests, we designed and fabricated a \mbox{high-sensitivity} \mbox{\textmu{N}-resolution} force sensor. Employing this sensor, we discovered that the tested \mbox{single-tail} propulsor generates relatively small \mbox{cycle-averaged} force values, with the best case being \mbox{$2.97$\,{\textmu}N}, which occurs at low actuation frequencies (\mbox{$\leq 2$\,Hz}); by comparison, we measured \mbox{cycle-averaged} force values of up to \mbox{$22.6$\,{\textmu}N} for the \mbox{dual-tail} propulsor. 

Interestingly, we also found that, at respectively \mbox{$0.45$\,mN} and \mbox{$0.61$\,mN}, the measured peak force values for the \mbox{single-tail} and \mbox{dual-tail} cases are very similar. Considering these data, we hypothesize that the \mbox{dual-tail} propulsor generates a significantly higher average thrust due to the interaction between the wake structures created by its two composing soft tails. This hypothesis is consistent with the finding that the \mbox{reactive-force} model presented in \mbox{Section\,\ref{SECTION02}} estimates force values similar to those measured for the \mbox{single-tail} case; in contrast, a noticeable mismatch between the modeled and experimentally obtained propulsion forces occurs in the \mbox{dual-tail} case. This discrepancy is most likely due to the omission of the convective term in the formulation of (\ref{EQ06}), which seems to be necessary to account for interaction between wake structures. In the near future, we will refine the \mbox{reactive-force} model presented in \mbox{Section\,\ref{SECTION02}} to account for \mbox{wake-structure} phenomena. Aligned with this objective, we will improve our understanding of the dynamics relating actuation output with thrust for locomotion, which is essential for developing advanced control schemes for \mbox{insect-scale} aquatic robots.

\bibliographystyle{IEEEtran}
\bibliography{references}

{\small
\section*{Appendix: Reactive-Force Model}
\vspace{-0.5ex}
\label{Appendix}
\noindent In this appendix, we review the fundamental ideas introduced by Lighthill in\cite{LighthillMJ1971} to model the total reactive force acting on a slender undulating tail. Here, we start by finding the material acceleration of a parcel of fluid at $\left(s,t\right)$ as
\begin{align}
\bs{a}(s,t) = \frac{D\bs{v}}{Dt} = a_{\ts{n}} \bs{u}_{\ts{n}} + a_{\ts{t}} \bs{u}_{\ts{t}},
\label{EQ10}
\end{align}
with 
\begin{align}
\begin{split}
a_{\ts{n}}(s,t) &= \frac{Dv_{\ts{n}}}{Dt} + v_{\ts{t}} \frac{\partial\theta}{\partial t} + v_{\ts{t}}^2 \frac{\partial \theta}{\partial s}, \\
a_{\ts{t}}(s,t) &= \frac{Dv_{\ts{t}}}{Dt} - v_{\ts{n}} \frac{\partial\theta}{\partial t} - v_{\ts{n}} v_{\ts{t}} \frac{\partial \theta}{\partial s},
\end{split}
\label{EQ11}
\end{align}
where $\theta(s,t)$ is the local tangent angle of the tail centerline from $\bs{b}_1$. Next, we invoke Newton's second and third laws to obtain the hydrodynamic force per unit length exerted by the fluid on the tail,
\begin{align}
\begin{split}
\hspace{-4ex}
\bs{f}(s,t) &= -m(s) \bs{a}(s,t) \\
&= -m(s) \left[\left( \frac{Dv_{\ts{n}}}{Dt} + v_{\ts{t}} \frac{\partial\theta}{\partial t} + v_{\ts{t}}^2 \frac{\partial \theta}{\partial s} \right) \bs{u}_{\ts{n}} \right.\\
&
\left. \hspace{8ex}
+\left( \frac{Dv_{\ts{t}}}{Dt} - v_{\ts{n}} \frac{\partial\theta}{\partial t} - v_{\ts{n}} v_{\ts{t}} \frac{\partial \theta}{\partial s} \right) \bs{u}_{\ts{t}} \right],
\end{split}
\label{EQ12}
\end{align}
where $m(s)$ is the added mass per unit length, which in this case is specified by (\ref{EQ04}). Additionally, we consider that\,(i)\,the motion of the tail's centerline is confined to the \mbox{$\bs{b}_1$--$\bs{b}_2$};\,(ii)\,$\bs{\mathcal{B}}$ does not rotate relative to $\bs{\mathcal{N}}$;\,(iii)\,the local curvature defined by $\theta(s,t)$, and its variations over $s$ and $t$ are small; and, following Lighthill's assumptions,\,(iv)\,reactive forces arise from the normal acceleration of the displaced fluid mass and act perpendicularly on the tail surface. Directly from these assumptions it follows that the reactive force per unit length is given by
\begin{align}
\bs{f}_{\ts{r}}(s,t) = -m(s) \bs{a}_{\ts{n}}(s,t) = -m(s) \frac{Dv_{\ts{n}}}{Dt} \bs{u}_{\ts{n}}(s,t), 
\label{EQ13}
\end{align}
in which the material derivative of the normal velocity component, $v_{\ts{n}}$, satisfies 
\begin{align}
\frac{Dv_{\ts{n}}}{Dt} = 
\frac{\partial v_{\ts{n}}}{\partial t} + \frac{ds}{dt} \frac{\partial v_{\ts{n}}}{\partial s}
=\frac{\partial v_{\ts{n}}}{\partial t} + v_{\ts{t}} \frac{\partial v_{\ts{n}}}{\partial s}.
\label{EQ14}
\end{align}
Therefore,
\begin{align}
\bs{f}_{\ts{r}}(s,t) = -m(s) \left( \frac{\partial v_{\ts{n}}}{\partial t} +  v_{\ts{t}} \frac{\partial v_{\ts{n}}}{\partial s} \right) \bs{u}_{\ts{n}}(s,t), 
\label{EQ15}
\end{align}
which allows us to compute the total instantaneous reactive force acting on the tail as specified by (\ref{EQ05}).
}

\end{document}